\PassOptionsToPackage{table, svgnames, dvipsnames}{xcolor}

\documentclass[runningheads]{llncs}
\usepackage{amsmath,amssymb,amsfonts}
\usepackage{algorithmic}
\usepackage{graphicx}
\usepackage{textcomp}
\usepackage{multirow}
\usepackage[short]{optidef}
\usepackage{xcolor}
\usepackage{multicol}
\usepackage{float}
\usepackage[colorinlistoftodos]{todonotes}

\usepackage{dblfloatfix}

\definecolor{indianred}{rgb}{0.8, 0.36, 0.36}

\usepackage{mathtools}
\usepackage[ruled,vlined]{algorithm2e}
\usepackage{caption}
\usepackage{subcaption}
\usepackage{booktabs}
\usepackage{array}

\newcommand{\printfnsymbol}[1]{%
  \textsuperscript{\@fnsymbol{#1}}%
}
\newcommand{\repeatthanks}{\textsuperscript{\thefootnote}}

\newcolumntype{C}[1]{>{\centering\let\newline\\\arraybackslash\hspace{0pt}}m{#1}}

\begin{document}
\title{Multimodal Neural Network For Demand Forecasting\\}
%
%
\author{Nitesh Kumar\thanks{Both the authors have made equal contribution towards the paper.} \and
Kumar Dheenadayalan\repeatthanks \and
Suprabath Reddy \and
Sumant Kulkarni}

\authorrunning{Kumar et al.}

\institute{Zenlabs, Zensar Technologies Ltd.}


%
\maketitle              
\begin{abstract}
	Demand forecasting applications have immensely benefited from the state-of-the-art Deep Learning methods used for time series forecasting. Traditional uni-modal models are predominantly seasonality driven which attempt to model the demand as a function of historic sales along with information on holidays and promotional events. However, accurate and robust sales forecasting calls for accommodating multiple other factors, such as natural calamities, pandemics, elections, etc., impacting the demand for products and product categories in general. We propose a multi-modal sales forecasting network that combines real-life events from \textit{news articles} with traditional data such as historical sales and holiday information. Further, we fuse information from general product trends published by Google trends. Empirical results show statistically significant improvements in the SMAPE error metric with an average improvement of \textbf{7.37\%} against the existing state-of-the-art sales forecasting techniques on a real-world supermarket dataset.

\end{abstract}

\section{Introduction}
Demand/Sales forecasting deals with predicting the number of goods and services that might be consumed under prevailing market conditions \cite{armstrong}. Forecast of demand can affect other aspects of the supply chain, like raw material procurement, human resource estimation, manufacturing, inventory management by retailers, etc. Hence, accurate forecasting enables businesses to be effective, reduce financial losses, and increase consumer satisfaction by meeting their availability requirements. Many visible events like social unrest, pandemic, economic recession, extreme weather activity which have an impact on the forecasting, cannot be easily quantified \cite{THOMASSEY2010}. Such events are mostly observed through multimodal data (text - news/policy articles, image - weather report, video - news recordings; time-series data like weather recordings) \cite{mansf}. Hence, it is essential to incorporate them as inputs for more accurate demand forecasting. 

The effect of an event can be positive, negative or neutral on the sale of a product category $(C)$. For example, weather-sensitive products like umbrellas, raincoats, bread, milk will see an increase in sales in specific weather conditions, and a similar pattern can be seen for air purifiers in California during forest fires \cite{californiaFire}. The event here is the forest fire, and the influence/effect of the event on sales is `+'ve.  The estimation of influence is an important step in incorporating unstructured text data for sales forecasting. To embed event-related information for sales forecasting, we use a Multimodal AI (MAI) model. Here, multiple independent models operate on different sources/modes of input that are eventually combined using fusion strategies like mid-fusion or late-fusion \cite{fbsound2} to make the final forecast. 

One of the major challenges in multimodal networks is the difficulty in combining different modalities to achieve a particular goal without facing generalization issues. Ideally, the addition of newer input sources should lead to more useful information for the forecasting task but training an end-to-end multimodal model is not as competitive as their unimodal counterparts in many cases \cite{unimodedis1}. The main causes for this as reported in \cite{fbmultimodal} are :
\begin{itemize}
	\item Multimodal models are often prone to overfitting due to the increased capacity of the model.
	\item Different modalities overfit and generalize at varying rates, so training them jointly with a single optimization strategy is sub-optimal.
\end{itemize}
Hence, it is important to optimally integrate the modalities (which are possibly independent networks) for robust forecasting. In our proposed model, we use historical sales data, google trends related to some of the product categories and news articles as inputs sources. To deal with the aforementioned issues of multimodal networks, we use the Overfitting to Generalization Ratio (OGR) \cite{fbmultimodal} as a criterion for weight updates that helps in alleviating the issues in a multimodal network and the same is covered in detail in Section \ref{sec:ogr}.\newline




Our contributions in this paper include:
\begin{itemize}
	\item Usage of multi-source and multimodal data (sales history, google trends, news articles and date) to build a forecasting engine that:
    \begin{itemize}
        \item shows statistically significant results compared to the existing state-of-the-art models like DeepGLO \cite{deepglo} and Filternet \cite{filternet}
        \item achieves higher forecast accuracy and better forecast bias compared to the existing literature
    \end{itemize}
	\item We show the importance of different modes of information based on the events occurring in a geographic region. Events and their impact encoded through news articles and their influence on estimating the sales numbers are demonstrated.
	\item We also show the advantage of monitoring (OGR) \cite{fbmultimodal} for successive weight updates that are necessary to ensure better generalization of multimodal networks.
\end{itemize}
\label{sec:intro}

The paper is divided into 5 sections:
\begin{itemize}
    \item The Problem Formulation - We describe the mathematical formulation of the forecasting problem
    \item Related Work – We cover the relevant literature for 3 important aspects in our approach, namely - (i) Deep Learning for time series, (ii) Representation learning from news articles and (iii) Multimodal AI.
    \item Data collection, pre-processing \& Feature Encoding – We cover the collection, processing and encoding process for 4 major sources of information used by our forecasting model. The input sources include  (i)sales data (ii)news articles (iii)google trends (iv)date
    \item The Multimodal Forecasting Model – We describe our Multimodal Forecasting model which is inspired by the Filternet architechture \cite{filternet}. This architechture has shown great promise in terms of performance and robustness when applied to different domains.
    \item Results – We present the key results and benchmarks against the uni-modal version of the problem. We follow up the results with a brief discussion section covering our key take-aways.
\end{itemize}


\section{The Problem Formulation}
\label{sec:formulation}
To formalize the sale forecasting problem, let  $S = (s^{p}_{t}: t \in T)$ and $G = (g^{q}_{t}: t \in T)$ be two time series consisting of daily historical sales of different product categories ($p$) and google search trends for trend category ($q$) respectively. Additionally, let $n_{t} \in N$ represent the set of news articles on day `t'. Given this dataset, we aim to develop a forecasting model that approximates \[ y^{p}_{t} = f(s^{p}_{t-1:t-w_{s}}, g^{q}_{t-1:t-w_{g}},\phi(n_{t-1:t-w_{n}}), \psi(t))\]

Here, the forecast for day `\textit{t}' is estimated by a fusion of inputs from multi-source (sales, google trends) and multimode (time-series and unstructured text like news) at various frequencies as seen in the function being estimated. $w_{s}$, $w_{g}$ and $w_{n}$ represents the window size indicating the number of past time steps to be used for sales data, google trends, and news articles respectively. $\phi$ and $\psi$ are encoders of news and date information respectively. Details of individual sources and modes of data are discussed in the subsequent subsections. With this setup, we present the details of sales forecasting model that is inspired by the Filternet architecture \cite{filternet} which is robust when applied on different domains. We begin the discussion of our architecture by describing the method used for news article representation.

\section{Related Work}
\label{sec:litsurvey}
Time series is a sequence of physical measurements over uniform/non-uniform time intervals. It has been of interest historically in multiple domains like climate modelling, retail, finance, to name a few. Literature on forecasting dates back a few decades, and the same has evolved from a statistical model to deep learning models. Traditional statistical models like Auto Regression (AR), Moving Average (MA), Auto Regression Integrated Moving Average (ARIMA), \cite{ARIMA},  Seasonal Autoregressive Integrated Moving Average (SARIMA) and modern statistical models like \textbf{T}rigonometric \textbf{B}ox-Cox \textbf{A}RMA \textbf{T}rend \textbf{S}easonal (TBATS) \cite{TBATS} and Prophet \cite{fbProphet} have shown varying degrees of performance for sales forecasting. Though statistical models are still popular and find applications in the field of economics, they may not be enough for broader application on sales dataset \cite{whyarimaSarima}. Forecasting at scale is better achieved by Prophet \cite{fbProphet}, which is robust to outliers and also consists of many knobs to tweak and adjust the forecast. Statistical models are mostly linear and are unable to deal with data that show asymmetric non-linear trends \cite{hydman}. Going beyond statistical models, deep learning in the past decade has advanced the research in forecasting by a big margin \cite{deeplearningSurvey}. According to \cite{bengioNNAutomation}, this popularity can be attributed to the ability of deep learning models to automate the feature engineering process and learn the underlying complex representations. 

\subsection{Deep Learning for Time-Series}

Deep Neural Networks(DNN) have achieved state-of-the-art results in computer vision and natural language processing. With a large enough training dataset,  DNN's can successfully capture historical trends from time-series data into a latent encoded representation. Different architectures have been used in literature and the most popular being sequence modelling methods like Recurrent Neural Networks and LSTMs. The temporal aspect of time-series and language modelling encourages many researchers to use LSTMs and their variants in the forecasting domain. Further, there has been the successful application of Convolutional Neural Networks (CNN) \cite{cnn}, initially designed for computer vision applications, where spatial information is extracted and encoded locally. Scaling down the CNNs from 2D to 1D allows using CNNs for time-series datasets, both in univariate and multivariate form \cite{1DCNN-survey}. Similar to spatial invariance of 2D CNNs, temporal invariance is also true in the 1D CNNs. 
There are significant improvements in the time-series classification domain. A popular architecture for classification is the Long Short Term Memory with Fully Connected Network (LSTM-FCN) \cite{lstmfcn}. \cite{lstmfcn} introduced alternate versions of this architecture with an attention mechanism. Another significant improvement was proposed in \cite{filternet}, where multiple layers of stacked 1D CNN or LSTM units were used. The stacking at different layers enables input length invariance which makes it suitable for variable-length time series. 

There are hybrid models where an ensemble of simple models produces much better results than high capacity models \cite{hybrid2020,DSF_SalesForecasting_2019}. 
DNNs can be sensitive to data preprocessing \cite{deepglo,deepar},  hence, hybrid methods which combine statistical and deep learning models have gained traction in the recent literature and are showing significant generalization performance. DeepGLO \cite{deepglo} is another important work that uses global information to aid forecasting in an ensemble fashion. The process of combining multiple models dealing with different input sources in a deep learning setting offers a significant advantage over existing statistical and deep learning techniques which is explored in this paper and demonstrated in Section \ref{sec:results}.

\subsection{Representation Learning for news articles}
The idea of incorporating information from news articles has been a well-explored area in the context of stock market predictions. Traditional machine learning models were explored in \cite{5}, using tree-based kernels with SVMs for the prediction task. Deep sequential models such as LSTMs and GRUs have been extensively used to model time series data. 

Encoding events extracted from news articles using Deep Learning Framework \cite{1,2,3,newsEmbeddings} and then using them for stock prediction is a generic theme across the literature. Predicting future stock prices by modelling transcripts of the earnings call was proposed in \cite{4}. A hybrid attention network (HAN) was introduced in \cite{newsEmbeddings} to classify stock increase/decrease based on news articles relates to the problem at hand. We adopt the HAN model to estimate the `+'ve/ `-'ve influence of news on sales of products.

\subsection{Multimodal AI}
The existing literature in multimodal AI deals with identical tasks like the Kinetics video dataset where video clips, audio, and optical flow are used as data modalities for action recognition \cite{kineticdataset,kinetic1}. Similarly, the mini-sports dataset consists of audio and video for sports classification \cite{minisport}. Other tasks like Visual-Q\&A \cite{fbvqa,fbvqa2}, sound localization \cite{fbsound2} in videos also fall in the identical tasks multimodal AI framework. All these problems majorly deal with sourcing data and modalities with correlated information directly dealing with the problem at hand. For instance, action recognition data will have both video and audio giving information about the action performed in the video. However, the uniqueness of sales forecasting is the need to extract relevant events that can affect the sales and associate these events with a subset of categories which adds to the complexity. There are very few news articles published daily that can directly impact the sales and identify as well as associating the news with product sales is a unique and challenging task that we address in our work.

\section{Data Collection, Pre-processing and Feature Encoding}


\subsection{Sales Data}
In our experiments, we used sales forecasting dataset published by a supermarket company, Corporación Favorita \cite{kaggle_ecuador_dataset}, which operates out of Ecuador. The daily sales number for each item is provided from 2013 to 2017 with each item belonging to 21 different product categories, but the data for several products are sparse. Hence, we aggregate the product category-wise sales and consider 20 product categories ($C_1$ - $C_{20}$) in our experiment, while omitting 1 category (LAWN AND GARDEN) due to extreme sparsity of sales numbers. A total of 1610 days’ data was aggregated for each of the 20 sales product categories of which 1300 instances were used for training and the rest were used for testing. This roughly amounts to 3 years of data for training and 1 one year data for testing.

\subsection{News Dataset}
\label{subsec:newsencoding}
We collect news articles from popular news websites that operate in Ecuador and other South American regions for the time-period for which we have the sales data. Overall, we collected more than $125$K+ articles from $3$ websites - USNews\footnote{https://www.usnews.com/topics/locations/ecuador}, El Commercio\footnote{https://elcomercio.pe/}, and BBC News\footnote{https://www.bbc.com/news}. Since a lot of these news articles were in native South American languages, we use Google Translate \cite{googletrans} to translate them into the English language. During this process, it was ensured that this extraction is not biased towards a particular category of news such as financial or political since we do not want to make apriori assumptions about what type of news might impact product sales. We release this dataset\footnote{https://www.kaggle.com/reviewerh/news-dataset} for future research purposes. Learning the representations from news articles consists of the following $3$ steps.

\begin{enumerate}
	\item $Keyword\ Generation$ -  Generate candidate keyword sets $S_1$ - $S_{20}$ for each category $C_1$ - $C_{20}$.
	\item $Relevancy\ Scoring$ - Associate news articles to one/more category(s) using a scoring function, enabling us to filter out noisy and non-informative news articles. 
	\item $News\ Encoding$ - We learn the embeddings of articles using the Hybrid Attention Network \cite{newsEmbeddings} which are later used by our multimodal network for sales forecasting.
\end{enumerate}

\subsubsection{Keyword generation:} 
\begin{figure}
    \centering
    \includegraphics[width=\textwidth]{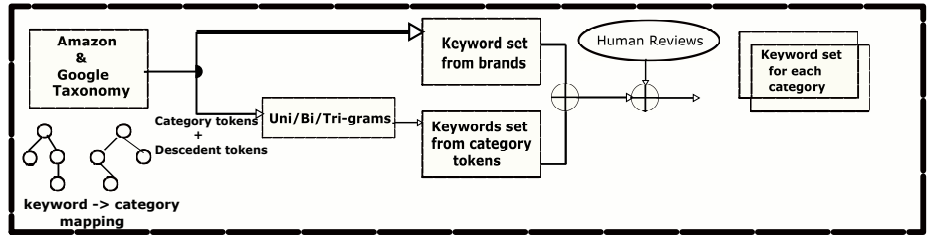}
    \caption{Keyword Generation}
    \label{fig:keyword}
\end{figure}

A list of candidate keywords are generated for each of the product categories as shown in Fig \ref{fig:keyword}, such that the news corpus can be queried with these keywords and later ranked according to their relevance towards each category. 
To accomplish this, firstly, we manually map each of the product categories ($C_1$ – $C_{20}$) to the appropriate entities in Amazon\footnote{http://jmcauley.ucsd.edu/data/amazon/} and Google taxonomy\footnote{https://www.google.com/basepages/producttype/taxonomy-with-ids.en-US.txt}. Secondly, we consider all sub-categories (descendants) of the mapped entities from both the taxonomies to create uni-gram, bi-gram, and tri-gram tokens for each product category. This served as the first set of candidate keywords - keywords coming from taxonomy entities. Further, we enriched the keyword set $S_i$ for category $C_i$ by extending the keyword set to all the brands associated with the	mapped entity in Amazon taxonomy. This gave us a larger keyword set so that we don’t miss out on news articles that might be relevant to our categories. A final round of manual pass was done to remove any generic keyword. By the end of this exercise, for each of the 20 categories $C_1$ – $C_{20}$, we had keyword sets $S_1$ – $S_{20}$. 

A sample of the keywords generated are shown in Table \ref{tab:keyword}
\begin{table}[h]
	\centering
	\caption{Sample keywords generated for subset of categories.}
	\label{tab:keyword}
		\resizebox{\linewidth}{!}{
		\begin{tabular}{ c|C{4cm}|C{4cm}|C{1.5cm} }
			\toprule
			$\text{Category Name}$ & $\text{Taxonomy Keywords}$ & $\text{Brand Keywords}$ & $\text{Set Index}$\\
			\midrule
			AUTOMOTIVE &   board lights, transmission filters, bellows kits ... pumps accessories, antilock & Fey, King Motor RC, Green Hill Graphics ... Raceceiver, ACOSUN, Pusher Intakes & $S_{1}$\\

			\vdots    & \vdots & \vdots &  \vdots\\
			BREAD BAKERY &   bakery, muffins, sandwich breads, $\ldots$,  fresh baked cookies, hamburger buns  & Tennenkoubo, Vitalicious ... Bantam Bagels , Home Pride, Townhouse,   &$S_4$\\

			\vdots    & \vdots & \vdots &  \vdots\\
			
			DAIRY & yogurt starter cultures, almond milk ... cotija, milk substitutes& Milk \& Co., Trumoo, Hormel Healthlabs ... Old Chatham Sheeping Herding &  $S_6$\\
			\vdots    & \vdots & \vdots &  \vdots\\
			
			\bottomrule
			
		\end{tabular}
	}
\end{table}

\subsubsection{Relevancy scoring:} 
\begin{figure}[ht]
	\centering
	\includegraphics[scale=.5]{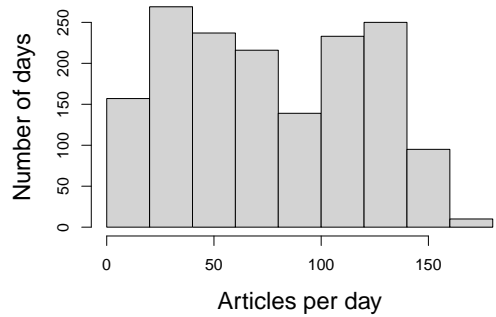} 
	\caption{Frequency of days and articles}
	\label{fig:art-freq}
\end{figure}

\begin{figure}
    \centering
    \includegraphics[width=\textwidth]{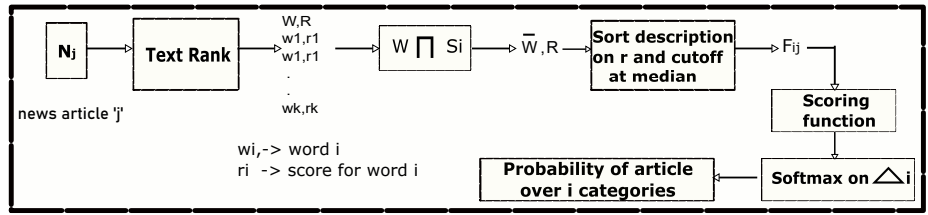}
    \caption{Keyword Generation}
    \label{fig:art2cat}
\end{figure}

Not all articles will carry information pertinent to the product categories under consideration. We term these articles as noisy articles. The proportion of noisy articles in the acquired news dataset is quite large compared to meaningful articles. The variability in the number of news articles per day is as shown in  Fig. \ref{fig:art-freq}
Although we see that there are $10$'s (even $100$'s) of news articles collected each day, upon manual exploration, we found that only a few articles per day are relevant to at least one of the $20$ categories.  This motivated us to design a relevancy module that would filter out top $k$ articles per day based on their relevancy to one or more categories and the flow for the same is presented in Fig \ref{fig:art2cat}. 
The motive is to assign a probability distribution over the $20$ categories for each news article if it is relevant to at least one category, else mark it as irrelevant. 

As a first step, we use TextRank \cite{7} to find the importance score for every word in a given news article $N_{j}$. For any category $C_{i}$, we find the keywords which are present in the article $N_{j}$ as well as in the keyword set $S_i$. Let this set have \textit{`l'} keywords, denoted by $F_{ij} = w_{1}, w_{2}, \ldots, w_{l},$ and let the TextRank scores of these keywords be denoted by $r_1, r_2 \ldots, r_l$. We further reduce the size of $F_{ij}$ by considering top $50\%$ keywords according to their descending order of  TextRank scores. We denote this new set by \[\hat{F}_{ij} = w_1, w_2, \ldots, w_h, \text{where } h=\frac{l}{2}\].

Now the relevancy score of an article $N_j$ towards $C_{i}$ is defined by 2 components -
\begin{itemize}
	\item How important is a keyword $w_k \in \hat{F}_{ij}$ to the category $C_{i}$. This is denoted by $\frac{1}{m_k}$, where $m$ is the number of categories that have the keyword $w_k$.
	\item How important is a keyword $w_k \in \hat{F}_{ij}$ to the article $N\textsubscript{j}$. This is given by \[ r_k \star \frac{1}{\text{rank}(w_k)}\]
	where $rank(w_k)$ is the index of sorted keywords according to TextRank score \textit{`r'}.
\end{itemize}

The combined score is given by \[\text{score}_{ij} = \sum_{k=1}^{h} \frac{1}{m_k} \star  r_k \star \frac{1}{\text{rank}(w_k)}\].
We then calculate this score for every $C_{i}$, $N_{j}$ pair. We review the top scores for each category to define a threshold $t_i$ above which the articles showed significant relevance towards category $C_{i}$. We then define \[\bigtriangleup_{ij} = \text{score}_{ij} - t_i\] as the actual relevance score between $C_{i}$ and $N_{j}$. Articles which had  $\bigtriangleup_{ij} \leq 0$ for every $C_{i}$ were deemed irrelevant. For articles which had ${\bigtriangleup}_{ij} > 0$ for at least one category were deemed relevant. 
At the end of this exercise, the maximum number of relevant articles ($n\textsubscript{articles}$) in a day (irrespective of the number of categories it was relevant to) turned out to be $5$. Thus, at most $5$ relevant articles per day were fed into the encoding module.

The relevance distribution over categories for an article $N_j$ was calculated as  \[
P_{ij} = \frac{e^{\bigtriangleup_{ij}}}{\sum_{\forall_{i}}e^{\bigtriangleup{ij}}}
\].


\subsubsection{News Encoding }

We generate targeted embeddings from relevant news articles such that they are able to explain rise/dip of sale values in the product categories. To accomplish this, we use the Hybrid Attention Network proposed in \cite{newsEmbeddings} and frame the problem as a classification task with a few modifications as follows -

\begin{itemize}
	\item The input fed to the network was a tensor of shape $n\textsubscript{articles} \star n\textsubscript{days} \star n\textsubscript{dim}$, where $n\textsubscript{articles}=5$ is the maximum number of relevant articles per day, $n\textsubscript{days}=7$ is the number of days in the sequence (same as $w_{n}$ defined in Section \ref{sec:formulation}), and $n\textsubscript{dim}=100$ is the dimension of doc2vec \cite{8} embedding for each of the articles. Doc2Vec was trained on a larger corpus of news articles.
	
	\item The output for the classification task had $40$ classes - $2$ binary variables for each of the $20$ product categories constructed as follows :
	
	\begin{minipage}{.45\textwidth}
		\centering
		\[
		{C\textsubscript{i}}\textsuperscript{+} = \begin{dcases*}
			1  & sales increased by $5\%$\\
			0 & otherwise
			\label{cincrease}
		\end{dcases*}
		\]
	\end{minipage}
	\begin{minipage}{.45\textwidth}
		\[
		{C\textsubscript{i}}\textsuperscript{-} = \begin{dcases*}
			1  & sales decreased by $5\%$\\
			0 & otherwise
			\label{cdecrease}
		\end{dcases*}
		\]
	\end{minipage}
	\\
	\item The bi-GRU layer was replaced with a unidirectional GRU.
	
	\item Batch normalization was applied to the outputs of the first attention layer.
	
\end{itemize}
We used the $64$ dimension vector from the second attention layer as the vector representation which encapsulates information present in all the relevant news articles over the past $7$ days. This embedding is used in the downstream forecasting model as described in Section \ref{sec:forecasting_model}


\subsection{Date Encoding}
\label{sec:dateencoding}
As the day, week, and month in a time-series data plays an important role in identifying cyclic patterns, an appropriate encoding of the same ensures better learning in a forecast model. It also aids in faster learning using any of the popular optimization algorithms. We use the cyclic feature encoding for $4$ pieces of information, namely, the day of the month, day of the year, week of the month, and month of the year using 
\[x_{sin} = \sin(\frac{2 * \pi * x}{\max(x)})\],
\[x_{cos} = \cos(\frac{2 * \pi * x}{\max(x)})\].
Maximum value of x in the equations is = \{($28$-$31$) days, ($365$-$366$) days, {$4$-$5$} weeks, $12$ months\} for the $4$ different date information that we want to encode. This results in a 8-dimensional embedding for each date. 
Our experiment setup uses the date encoding only for the date for which we forecast rather than the traditional time-series forecasting approach where every sales data is associated with a date. 


\subsection{Google Trends}
\label{subsec:googletrends}
Use of google trends\footnote{http://trends.google.com/trends} and search keywords to better predict the interest of consumers in products has been of interest in the past. A relationship between google search keywords and the spike in influenza in the general population was established in \cite{googletrends1}. The search volume of a particular product can be associated with the interest in the product and the same can be translated to the sales volume as well. \cite{boxoffice} presented a similar line of work for box-office revenue. For the purpose of sales forecasting, not all product categories in the sales dataset have the corresponding google trends available. We manually skimmed through the available categories and identified $12$ google trend categories between 2013 to 2017 that were either exactly matching or closely related to the categories present in the sales data for the Ecuador region. The list of trend categories are - {Food \& Drink, Internet \& Telecom, Computers \& Electronics, Business \& Industrial, Shopping, Books \& Literature, Hobbies \& Leisure, Autos \& Vehicles, Home \& Garden, Arts \& Entertainment, Beauty \& Fitness, Health}.

\subsubsection{Correlation Analysis - } Since we have $12$ different trend categories ($q$) and $20$ product categories  ($p$), we perform a pairwise Pearson's correlation for each pair of trend time series and product sales time series, and pick the $q$ which has the maximum correlation with ($p$) above the threshold of $0.4$. The most correlated $q$ will be fed to the Trend-network that will later be part of the mid-fusion (described in Section \ref{sec:forecasting_model}). The purpose of setting a high threshold is to use only useful google trends for our task for sales forecasting. Consequently, not all product categories will have an associated Google trend. All the product categories which are associated with a Google trend category are marked in (*) in Section \ref{sec:results}.


\section{The Multimodal Forecasting Model }
\label{sec:forecasting_model}
\begin{figure*}
	\centering
	\includegraphics[ width=.95\textwidth]{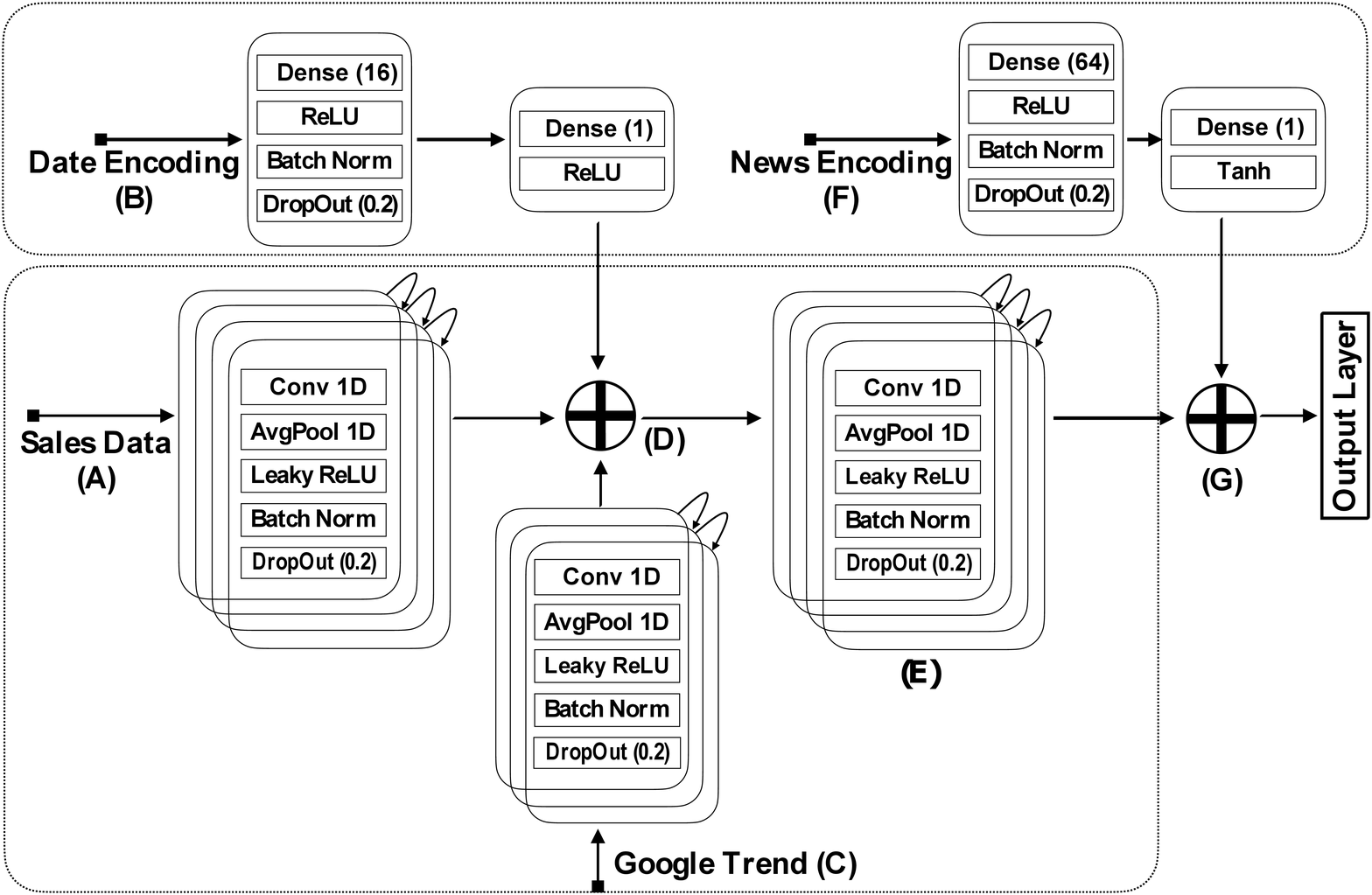} 
	\caption{Multimodal network architecture for sale forecasting.}
	\label{fig:mmn-block}
\end{figure*}

The overall architechture of our network is shown in Fig \ref{fig:mmn-block}. The news and date embeddings obtained in Section \ref{subsec:newsencoding} and Section \ref{sec:dateencoding} respectively are each fed into a fully connected layer with one output node. This output node acts as a gating unit to control the flow of information from these modalities into the forecasting system, thus enabling the network to learn if the modality needs to be used for forecasts in a specific period. 
In order to model the time series data coming from Sales data and the associated Google trend (if any), we draw inspiration from Filternet \cite{filternet} and use the idea of stacking Conv 1D blocks. The window length/sequence length for Sales and Google Trend data as described in Section \ref{sec:formulation} are $w_s = 30$ and $w_g = 30$ respectively. We have the first stack of Conv1D layers for the Sales time series (A) and another stack for the associated Google trend (C). Each of these stacks have 4 Conv1D blocks and each Conv1D block has kernel sizes varying from 1-4, thus generating 4 feature maps which are Average Pooled and followed by appropriate activation, batch-norm and dropout layers. The output of these two stacks are concatenated along with the output from the gated unit of Date Encoding as a mid-fusion strategy, as shown in step (D) of Fig \ref{fig:mmn-block}. The concatenated output is then passed through another stacking unit (E) of similar structure and the output of this stack is concatenated with the output of the gated unit of News Encoding as a part of our late-fusion strategy, as shown in step (G). This vector is then passed through the output layer which uses linear activation to generate a single value that serves as the regression output/forecast for the given inputs. 
Although filternet offers a choice of using either a stack of LSTM or Conv1D blocks, our experiments performed best with Conv1D blocks. All the Conv1D blocks use a leaky ReLU activation function and the fully
connected layers use the ReLU activation function. This apart, all the layers also
have batch normalization and dropout enabled. `L2' regularization is used to ensure that the model capacity is limited and the traditional problems of multi-modal networks as listed in Section \ref{sec:intro} are handled to some extent. We train this architecture separately for each of the 20 product categories - one for each category.



\subsubsection{Overfitting-Generalization-Ratio (OGR)}
\label{sec:ogr}
OGR, which is the ratio of overfitting ($\bigtriangleup O$) to generalization ($\bigtriangleup G$) was first introduced in \cite{fbmultimodal}, and used as a means of blending gradients with successful application in video classification. A rough estimation of  overfitting and generalization are shown in Equation \ref{eqn:overfit} and \ref{eqn:generalize} where $e$ refers to current epoch.
\begin{equation}
	\label{eqn:overfit}
	\bigtriangleup O	 = (\text{val\_loss}_{\text{e}} - 
	\text{train\_loss}_{\text{e}}) - \\
	(\text{val\_loss}_{(\text{e-1})} - 
	\text{train\_loss}_{(\text{e-1})})
\end{equation}
\begin{equation}
	\label{eqn:generalize}
	\bigtriangleup G	 = (\text{val\_loss}_{e}-\text{val\_loss}_{e-1})
\end{equation}
Intuition is that a small value of OGR indicates that the model is learning well with good generalization and less overfitting. It can also be interpreted as the quality of learning between epochs in the training phase. A similar idea is used in our work for successive weight updates. Instead of using OGR for gradient blending, we compare the OGR values in successive epochs in the training procedure and update the weights only if the OGR is less than a small number ($10^{-2}$).
We implement this simple yet effective check of measuring the improvements in generalization as well as overfitting for the current epoch with the previous epoch as the reference. This is contrary to the popular way of weight updates where successive epochs' training loss or validation loss is used. Using OGR, we ensure that the multimodal network does not get into the overfitting zone and ensure optimal updates only when the generalization of the model occurs with every weight update.


\section{Results}

\label{sec:data}

In the network shown in Fig \ref{fig:mmn-block}, we used Adam optimizer with 'MAE'  as the loss metric. The experiments were run for $200$ epochs, and one model per category was saved for testing. Window sizes (sequence length) for each input source (as described in Section \ref{sec:formulation}) are $w_s = 30$, $w_g = 30$ and $w_n = 7$. The models provided in \cite{lstmfcn}, \cite{deepglo}, and \cite{filternet} were used to benchmark LSTM-FCN, DeepGLO and Filternet respectively.
During testing, the first step is to gather and generate the encoding for the $3$ sets of inputs. These encodings then serve as training inputs to the multimodal network and generate the forecast. We have an option to generate results for \{1,7\} days and the results daily and weekly forecast are discussed next.


The experiment setup is consistent for all benchmark models. We used the dataset presented in Section \ref{sec:data}. In the first phase, we compare the results generated by the state-of-the-art deep learning models along with our proposed model. We also test the statistical significance of the model using the \textit{Wilcoxon signed-rank test}. It is a robust non-parametric statistical test used to compare two machine learning models across multiple datasets. Although the literature talks about its application in a classifier setting, there is no restriction in the application of the same in a forecasting model as well as all the deep learning models considered convert the data to a supervised setting.

	\subsubsection{Wilcoxon Signed-Ranks Test:}
	For statistical comparison of machine learning algorithms, research surveys have suggested the use of Wilcoxon Signed-Ranks Test as the most suitable statistical test when compared to other tests like paired t-test, averaging over multiple runs, etc., \cite{journal-Demsar-2006-mltest}. The Wilcoxon signed-ranks test \cite{journal-Demsar-2006-mltest} is a non-parametric equivalent to the paired t-test, used to rank the difference in the performance of two models in a multi-dataset scenario. 
	Since we train one model per product category, we have $20$ datasets across which the proposed model is tested. Keeping the model architecture constant, and treating each of the product categories as an independent dataset, we can use the signed-rank test to conclude if the results are statistically significant or not.

\subsection{Discussion}
\label{sec:results}

We initially started our experimentation with statistical models like SARIMAX, TABTS, FB-Prophet but none of them gave satisfactory results that were comparable with the deep learning models. Hence, the results presented here consist of the benchmarked deep learning approaches alone. The experiments were run over $10$ different iterations and the SMAPE mean along with standard deviation for each product category are presented in Table \ref{tab:baseline} for daily forecast. The results for our proposed multimodal network were statistically significant and the same was validated using the signed-rank test. 
We define the following hypothesis: 
	\begin{itemize}
		\item ${\displaystyle H_0}:$ there isn't any significant difference in SMAPE of the proposed and benchmark models, and the difference between each pair of SMAPE values follows a symmetric distribution around zero. 
		\item ${\displaystyle H_1}:$ there is a significant difference in SMAPE, and the difference between the pairs is not normally distributed around zero.
	\end{itemize}  
For an $\alpha = 0.5$ and $n = 20$, the signed-rank test for the best model in literature, i.e., filternet and the proposed multimodal network saw the value of z at $-3.7333$ and the p-value is $0.0002$. Owing to the low p-value, we reject the null hypothesis (both models are identical). The results for the proposed model without OGR are shown in column $5$ of Table \ref{tab:baseline}. 

Two key takeaways from this experiment were: 
	\begin{itemize}
		\item Results of our Multimodal Forecasting model \textit{without OGR} are competitive and comparable with the Filternet model. However, the statistical significance cannot be established as there are nearly equal proportions of product categories where either of them perform better than the other.
		\item The variance in the results obtained without OGR is relatively high. \textit{However, enabling OGR ensures that the model performs well across the dataset. }
	\end{itemize}  

\begin{table}[]
	\centering
	\caption{Daily forecast SMAPE results for all benchmarked models and Proposed architecture with and without OGR}
	\label{tab:baseline}
	\resizebox{0.9\linewidth}{!}{%
		\begin{tabular}{@{}lccccm{2cm}@{}}
			\toprule
			Name &
			DeepGLO  &
			LSTM-FCN  &
			Filternet  &
			Multimodal Network (with OGR)  &
			Multimodal Network (Non-OGR)  \\ \midrule
			$\text{\textbf{AUTOMOTIVE}}^{*}$       &   18.92 $\displaystyle \pm$ 4.26 \quad&\quad 14.28 $\displaystyle \pm$ 1.00 \quad\quad& 12.93 $\displaystyle \pm$ 0.43 \quad& \cellcolor[HTML]{FFCC67}12.17 $\displaystyle \pm$ 0.08 \quad& 12.41 $\displaystyle \pm$ 1.71 \\
			$\text{\textbf{BEAUTY}}^{*}$           & 20.94 $\displaystyle \pm$ 3.64 \quad& 23.36 $\displaystyle \pm$ 1.05 \quad& 15.55 $\displaystyle \pm$ 0.36 \quad& \cellcolor[HTML]{FFCC67}14.25 $\displaystyle \pm$ 0.16 \quad& 18.63 $\displaystyle \pm$ 4.98 \\
			$\text{\textbf{BEVERAGES}}^{*}$        & 17.05 $\displaystyle \pm$ 4.79 \quad& 15.26 $\displaystyle \pm$ 1.67 \quad& 10.79 $\displaystyle \pm$ 0.51 \quad& \cellcolor[HTML]{FFCC67}10.47 $\displaystyle \pm$ 0.82 \quad& 11.31 $\displaystyle \pm$ 2.78 \\
			$\text{\textbf{BREAD/BAKERY}}^{*}$     & 16.80 $\displaystyle \pm$ 5.40 \quad& 10.82 $\displaystyle \pm$ 0.64 \quad& 8.21  $\displaystyle \pm$ 0.19 \quad& \cellcolor[HTML]{FFCC67}7.08  $\displaystyle \pm$ 0.17 \quad& 7.69  $\displaystyle \pm$ 1.17 \\
			CLEANING         & 17.83 $\displaystyle \pm$ 4.96 \quad& 14.01 $\displaystyle \pm$ 0.70 \quad& 11.18 $\displaystyle \pm$ 0.38 \quad& \cellcolor[HTML]{FFCC67}9.97  $\displaystyle \pm$ 0.20 \quad& 10.45 $\displaystyle \pm$ 0.74 \\
			$\text{\textbf{DAIRY}}^{*}$            & 17.50 $\displaystyle \pm$ 4.76 \quad& 14.76 $\displaystyle \pm$ 1.33 \quad& 9.82  $\displaystyle \pm$ 0.30 \quad& \cellcolor[HTML]{FFCC67}9.04  $\displaystyle \pm$ 0.14 \quad& 9.47  $\displaystyle \pm$ 0.91 \\
			DELI             & 17.50 $\displaystyle \pm$ 4.84 \quad& 13.56 $\displaystyle \pm$ 0.88 \quad& 10.27 $\displaystyle \pm$ 0.30 \quad& \cellcolor[HTML]{FFCC67}8.71  $\displaystyle \pm$ 0.13 \quad& 10.30 $\displaystyle \pm$ 1.85 \\
			EGGS             & 19.02 $\displaystyle \pm$ 4.32 \quad& 16.19 $\displaystyle \pm$ 0.96 \quad& 11.16 $\displaystyle \pm$ 0.46 \quad& \cellcolor[HTML]{FFCC67}10.29 $\displaystyle \pm$ 0.40 \quad& 10.60 $\displaystyle \pm$ 0.86 \\
			$\text{\textbf{FROZEN FOOD}}^{*}$     & 22.85 $\displaystyle \pm$ 5.50 \quad& 19.06 $\displaystyle \pm$ 1.36 \quad& \cellcolor[HTML]{FFCC67}12.39 $\displaystyle \pm$ 0.43 \quad& 12.66 $\displaystyle \pm$ 1.01 \quad& 13.33 $\displaystyle \pm$ 2.08 \\
			GROCERY          & 17.23 $\displaystyle \pm$ 5.18 \quad& 15.07 $\displaystyle \pm$ 0.89 \quad& 10.35 $\displaystyle \pm$ 0.42 \quad& \cellcolor[HTML]{FFCC67}9.17  $\displaystyle \pm$ 0.30 \quad& 9.59  $\displaystyle \pm$ 0.85 \\
			HARDWARE         & 21.20 $\displaystyle \pm$ 3.39 \quad& 19.27 $\displaystyle \pm$ 1.04 \quad& 15.96 $\displaystyle \pm$ 0.16 \quad& \cellcolor[HTML]{FFCC67}15.90 $\displaystyle \pm$ 0.14 \quad& 18.90 $\displaystyle \pm$ 5.66 \\
			HOME APPLIANCES  & 30.11 $\displaystyle \pm$ 2.87 \quad& 33.10 $\displaystyle \pm$ 1.21 \quad& \cellcolor[HTML]{FFCC67}25.93 $\displaystyle \pm$ 0.17 \quad& 26.02 $\displaystyle \pm$ 0.13 \quad& 27.43 $\displaystyle \pm$ 4.02 \\
			LINGERIE         & 18.96 $\displaystyle \pm$ 4.57 \quad& 14.67 $\displaystyle \pm$ 1.11 \quad& 12.57 $\displaystyle \pm$ 0.17 \quad& \cellcolor[HTML]{FFCC67}11.34 $\displaystyle \pm$ 0.05 \quad& 12.01 $\displaystyle \pm$ 1.75 \\
			LIQUOR,WINE,BEER & 28.86 $\displaystyle \pm$ 4.04 \quad& 30.04 $\displaystyle \pm$ 3.03 \quad& \cellcolor[HTML]{FFCC67}21.09 $\displaystyle \pm$ 0.41 \quad& 21.99 $\displaystyle \pm$ 0.50 \quad& 22.81 $\displaystyle \pm$ 2.87 \\
			MEATS            & 18.26 $\displaystyle \pm$ 4.56 \quad& 14.21 $\displaystyle \pm$ 1.25 \quad& 9.70  $\displaystyle \pm$ 0.21 \quad& \cellcolor[HTML]{FFCC67}9.06  $\displaystyle \pm$ 0.17 \quad& 9.59  $\displaystyle \pm$ 1.10 \\
			PERSONAL CARE    & 18.63 $\displaystyle \pm$ 4.27 \quad& 14.87 $\displaystyle \pm$ 0.47 \quad& 12.26 $\displaystyle \pm$ 0.33 \quad& \cellcolor[HTML]{FFCC67}10.80 $\displaystyle \pm$ 0.09 \quad& 11.70 $\displaystyle \pm$ 1.82 \\
			POULTRY          & 18.87 $\displaystyle \pm$ 3.91 \quad& 15.25 $\displaystyle \pm$ 0.90 \quad& 11.29 $\displaystyle \pm$ 0.88 \quad& \cellcolor[HTML]{FFCC67}9.78  $\displaystyle \pm$ 0.17 \quad& 10.49 $\displaystyle \pm$ 1.53 \\
			PREPARED FOODS   & 17.09 $\displaystyle \pm$ 5.90 \quad& 11.73 $\displaystyle \pm$ 0.56 \quad& 8.45  $\displaystyle \pm$ 0.29 \quad& \cellcolor[HTML]{FFCC67}7.37  $\displaystyle \pm$ 0.10 \quad& 8.15  $\displaystyle \pm$ 1.10 \\
			$\text{\textbf{PRODUCE}}^{*}$          & 19.86 $\displaystyle \pm$ 4.31 \quad& 12.07 $\displaystyle \pm$ 1.63 \quad& 10.94 $\displaystyle \pm$ 1.02 \quad& \cellcolor[HTML]{FFCC67}10.23 $\displaystyle \pm$ 0.58 \quad& 10.95 $\displaystyle \pm$ 1.70 \\
			SEAFOOD          & 20.17 $\displaystyle \pm$ 4.24 \quad& 16.79 $\displaystyle \pm$ 1.26 \quad& 12.65 $\displaystyle \pm$ 0.41 \quad& \cellcolor[HTML]{FFCC67}11.55 $\displaystyle \pm$ 0.05 \quad& 12.30 $\displaystyle \pm$ 1.15 \\ \bottomrule
		\end{tabular}%
	}
\end{table}


\begin{table}[h]
	\centering
	\caption{Weekly results for entire dataset along with results during a major event}
	\label{tab:weekly-with-events}
	\resizebox{0.9\linewidth}{!}{%
		\begin{tabular}{@{}lcc|cc@{}}
			\toprule
			Name &
			Filternet (weekly) &
			Multimodal (weekly) & Filternet(event) & Multimodal Network(event) \\ \midrule
			$\text{\textbf{AUTOMOTIVE}}^{*}$        & 7.73  $\displaystyle \pm$  0.36 & \cellcolor[HTML]{FFCC67}6.82  $\displaystyle \pm$  0.39 & 5.71          $\displaystyle \pm$ 0.61         & \cellcolor[HTML]{67FD9A}5.20                   $\displaystyle \pm$ 1.71            \\
			$\text{\textbf{BEAUTY}}^{*}$           & 10.65 $\displaystyle \pm$  0.41 & \cellcolor[HTML]{FFCC67}9.92  $\displaystyle \pm$  1.14 & 7.28          $\displaystyle \pm$ 0.47         & \cellcolor[HTML]{67FD9A}6.50                   $\displaystyle \pm$ 1.05                \\
			$\text{\textbf{BEVERAGES}}^{*}$        & 7.93 $\displaystyle \pm$  1.08 & \cellcolor[HTML]{FFCC67}7.17 $\displaystyle \pm$  0.72  & 9.04          $\displaystyle \pm$ 1.08         & \cellcolor[HTML]{67FD9A}6.75                   $\displaystyle \pm$ 0.98 \\
			$\text{\textbf{BREAD/BAKERY}}^{*}$     & 4.58  $\displaystyle \pm$  0.42 & \cellcolor[HTML]{FFCC67}3.93  $\displaystyle \pm$  0.36 & 3.75          $\displaystyle \pm$ 0.63         & \cellcolor[HTML]{67FD9A}2.71                   $\displaystyle \pm$ 0.68 \\
			CLEANING         & 8.96  $\displaystyle \pm$  0.69 & \cellcolor[HTML]{FFCC67}7.74  $\displaystyle \pm$  0.33 & 7.59          $\displaystyle \pm$ 1.60         & \cellcolor[HTML]{67FD9A}5.91                   $\displaystyle \pm$ 1.02 \\
			$\text{\textbf{DAIRY}}^{*}$            & 5.27  $\displaystyle \pm$  0.90 & \cellcolor[HTML]{FFCC67}5.09  $\displaystyle \pm$  0.74 & \cellcolor[HTML]{67FD9A}5.83          $\displaystyle \pm$ 1.12         & 6.35                   $\displaystyle \pm$ 0.73\\
			DELI             & \cellcolor[HTML]{FFCC67}4.61 $\displaystyle \pm$ 0.26 & 4.63 $\displaystyle \pm$ 0.18 & 3.79          $\displaystyle \pm$ 0.75         & \cellcolor[HTML]{67FD9A}3.38                   $\displaystyle \pm$ 0.65  \\
			EGGS             & 5.15 $\displaystyle \pm$ 0.21 & \cellcolor[HTML]{FFCC67}3.80 $\displaystyle \pm$ 0.20 & 5.51          $\displaystyle \pm$ 0.60         & \cellcolor[HTML]{67FD9A}5.05                   $\displaystyle \pm$ 0.71   \\
			$\text{\textbf{FROZEN FOOD}}^{*}$     & \cellcolor[HTML]{FFCC67}12.69$\displaystyle \pm$ 2.63 & 12.88$\displaystyle \pm$ 2.04 & 27.64         $\displaystyle \pm$ 11.25        & \cellcolor[HTML]{67FD9A}22.67                  $\displaystyle \pm$ 5.21 \\
			GROCERY          & 9.97 $\displaystyle \pm$ 1.35 & \cellcolor[HTML]{FFCC67}8.97 $\displaystyle \pm$ 0.68 & 17.42         $\displaystyle \pm$ 1.81         & \cellcolor[HTML]{67FD9A}14.25                  $\displaystyle \pm$ 1.77 \\
			HARDWARE         & 9.05 $\displaystyle \pm$ 0.48 & \cellcolor[HTML]{FFCC67}8.14 $\displaystyle \pm$ 0.87 & 4.20          $\displaystyle \pm$ 0.76         & \cellcolor[HTML]{67FD9A}2.05                   $\displaystyle \pm$ 1.47\\
			HOME APPLIANCES  & \cellcolor[HTML]{FFCC67}10.48$\displaystyle \pm$ 1.38 & 11.09$\displaystyle \pm$ 1.09 & 29.26         $\displaystyle \pm$ 6.28         & \cellcolor[HTML]{67FD9A}26.73                  $\displaystyle \pm$ 4.41  \\
			LINGERIE         & 6.23 $\displaystyle \pm$ 0.45 & \cellcolor[HTML]{FFCC67}5.54 $\displaystyle \pm$ 0.58 & 4.25          $\displaystyle \pm$ 1.41         & \cellcolor[HTML]{67FD9A}3.09                   $\displaystyle \pm$ 1.78   \\
			LIQUOR WINE BEER & 14.59$\displaystyle \pm$ 1.20 & \cellcolor[HTML]{FFCC67}13.16$\displaystyle \pm$ 0.26 & 16.20         $\displaystyle \pm$ 4.45         & 15.26                  $\displaystyle \pm$ 2.53 \\
			MEATS            & 6.82 $\displaystyle \pm$ 0.38 & \cellcolor[HTML]{FFCC67}5.48 $\displaystyle \pm$ 0.56 & 4.07          $\displaystyle \pm$ 0.33         & \cellcolor[HTML]{67FD9A}3.06                   $\displaystyle \pm$ 0.44  \\
			PERSONAL CARE    & 11.45$\displaystyle \pm$ 0.74 & 7.88 $\displaystyle \pm$ 0.19  & 9.04          $\displaystyle \pm$ 0.98         & \cellcolor[HTML]{67FD9A}6.79                   $\displaystyle \pm$ 2.09  \\
			POULTRY          & 6.20 $\displaystyle \pm$ 0.63 & \cellcolor[HTML]{FFCC67}4.47 $\displaystyle \pm$ 0.37 & 4.13          $\displaystyle \pm$ 1.02         & \cellcolor[HTML]{67FD9A}2.64                   $\displaystyle \pm$ 0.62  \\
			PREPARED FOODS   & 4.59 $\displaystyle \pm$ 0.44 & \cellcolor[HTML]{FFCC67}4.09 $\displaystyle \pm$ 0.18 & 6.45          $\displaystyle \pm$ 0.61         & \cellcolor[HTML]{67FD9A}5.59                   $\displaystyle \pm$ 0.60  \\
			$\text{\textbf{PRODUCE}}^{*}$          & \cellcolor[HTML]{FFCC67}3.31 $\displaystyle \pm$ 0.56 & 3.64 $\displaystyle \pm$ 0.17 & \cellcolor[HTML]{67FD9A}2.31          $\displaystyle \pm$ 0.37         & 2.42                   $\displaystyle \pm$ 0.32  \\
			SEAFOOD          & 6.39 $\displaystyle \pm$ 0.39 & \cellcolor[HTML]{FFCC67}5.89 $\displaystyle \pm$ 0.34 & \cellcolor[HTML]{67FD9A}0.95          $\displaystyle \pm$ 0.66         & 2.84                   $\displaystyle \pm$ 0.38 \\ \bottomrule
		\end{tabular}
	}
\end{table}

The weekly forecast is also statistically significant (z-value is $-3.9199$ \& p-value is $0.00008$), and the results for the same is captured in Table \ref{tab:weekly-with-events} in columns $1$ \& $2$. The weekly results were generated for $40$ weeks in $2017$. To understand the use of modalities in identifying demand shifts, we focus on the forecast results obtained for product categories during a major flood that affected a major part of South America between $3\textsuperscript{rd}$ week of March $2017$ to $2\textsuperscript{nd}$ week of April $2017$  and the same is well captured in all the news articles as well as trends. The results for Filternet and the proposed model during this specific period are shown in Table \ref{tab:weekly-with-events} columns $3$ \& $4$. In line with the comparison, w.r.t. overall forecast accuracy, the SMAPE values during the event are reasonably good with additional information having a positive impact on $17$ of the $20$ product categories. For a similar statistical test setting of $\alpha=0.05$ and $n=20$, the value of z is $-3.1733$ the p-value is $0.00152$. Thus, validating the significance of the improved forecasts during the event.



\subsection{DeepShap Feature Importance}
To understand the importance of using additional sources and modes of data, we use DeepShap \cite{deepshap}, an open-source package for explainability which estimates shapeley values to understand the contribution of features in model prediction. 

For interpretation purpose, it must be noted that the larger the absolute Shapley values of a feature, the greater is its influence on the forecast made by the model. The x-axis represents the Shapley values and the y-axis represents the features. The jittered representation for each feature gives an indication of the distribution of Shapley values for each feature. The pink color represents a positive influence on the forecast, and blue is used for negative impacts. 


\begin{figure}
	\centering
		\includegraphics[width=.7\textwidth]{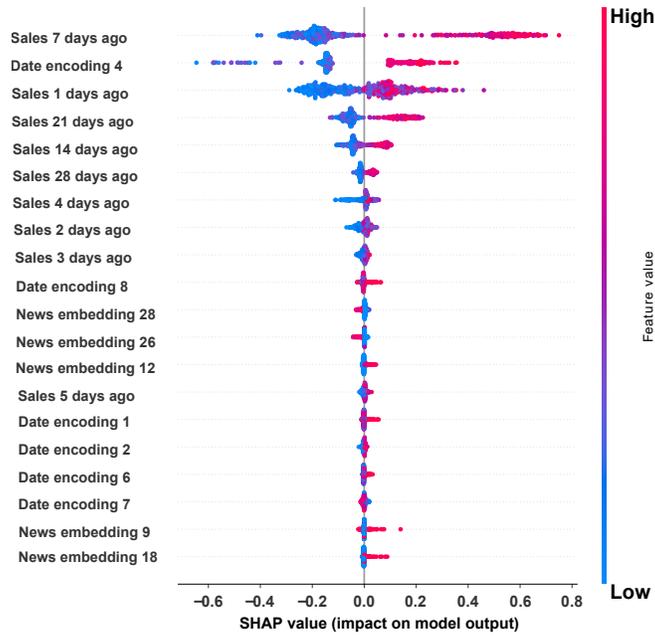}
		\caption{Feature importance when trained on a subset of data with events.}
		\label{fig:shapley-1}
		\end{figure}
\begin{figure}

		\centering
		\includegraphics[width=.7\textwidth]{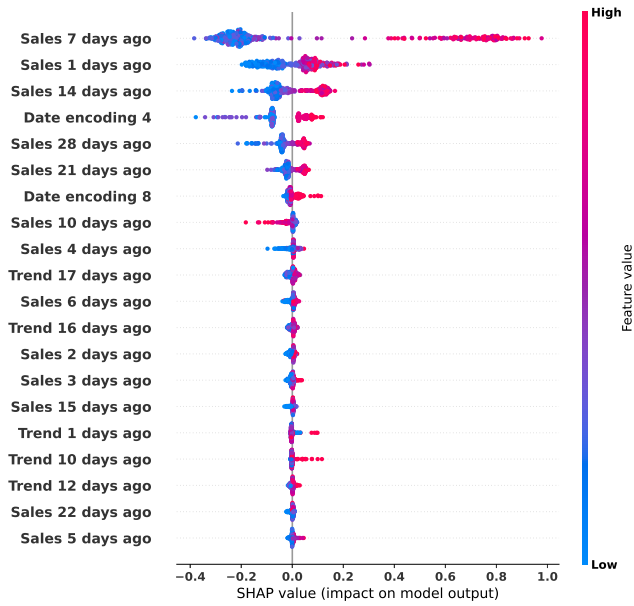}
		\caption{Feature importance when trained on a subset of data without events.}
		\label{fig:shapley-2}
\end{figure}
The summary plot shown in Fig \ref{fig:shapley-1} \& \ref{fig:shapley-2} shows the feature importance for two models trained on data with and without major events respectively. 
In Fig \ref{fig:shapley-1} the model was trained with data collected during major events (i.e., floods and earthquake during $2015$ to $2016$) and we see $5$ features related to news are influencing the predictions. The extent of the impact is relatively small as Shapley values are averaged over the entire data. However, the influence of events is still positive, as seen by the pink-colored representation of Shapley values.
In Fig \ref{fig:shapley-2}, the summary plot is shown for a model trained using data without any significant event (i.e., $2013$ - $2014$). We do not observe any news encoding feature but $5$ trend-based features appear to have a positive impact on the forecast. The top feature identified in both the scenarios discussed is the sales data 7 days in the history which shows that the weekly trends have the most impact followed date and the previous day's sales numbers (one day lookback). 


\section{Conclusion}
In this work, we have proposed a novel approach to automate the process of encoding major/minor events that directly affect the sales of commodity categories by collating information from various sources namely, online search trends and news articles published daily. We have achieved improved forecast accuracy with the multimodal architecture under general conditions and significant improvements under event-specific conditions. 

Since the multimodal network provides scope for integrating additional modalities, exploring more modalities is also an option with the caveat of carefully preventing overfitting. Weather data is another source that can be included right away depending on the geography, which is a useful source of information. As part of the future work, exploring better models rather than deep networks for learning other sources of data, capable of capturing the intricate details would be a good step in making multimodal network architecture more useful for applications.



\bibliographystyle{unsrt}
{
 \bibliography{resources/ecml-pkdd.bib}
}
\end{document}